%% file: example.tex
\documentclass{article}
\usepackage{amssymb}
\usepackage{multirow}
\usepackage{booktabs} 
\usepackage{amsmath}  
\usepackage{graphicx}
\usepackage{capt-of}
\usepackage{algorithm}
\usepackage{algpseudocode}
\usepackage{etoc}
\usepackage{gensymb}
\usepackage{makecell}
\usepackage{wrapfig}
\usepackage{cuted}
\usepackage{xspace}
\usepackage[accsupp]{axessibility}
\usepackage[final]{corl_2025}
\newcommand{\name}[0]{\textsc{Dexplore}\xspace}
\title{\name: Scalable Neural Control for Dexterous Manipulation from Reference‑Scoped Exploration}

\author{Sirui Xu$^{1,2}$\thanks{Work done during a part-time internship at NVIDIA Research.} \quad Yu-Wei Chao$^{2}$ \quad Liuyu Bian$^{1}$ \quad Arsalan Mousavian$^{2}$ 
\\
\textbf{Yu-Xiong Wang}$^{1\dag}$ \quad
\textbf{Liang-Yan Gui}$^{1\dag}$ \quad \textbf{Wei Yang}$^{2\dag}$\\
$^{1}$ University of Illinois Urbana-Champaign \quad $^{2}$ NVIDIA\\
$^{\dag}$ Equal Advising\\
{\small \url{https://sirui-xu.github.io/dexplore}}}

\begin{document}
\maketitle

\begin{figure}[H]\centering

\vspace{-2em}
\includegraphics[width=\textwidth]{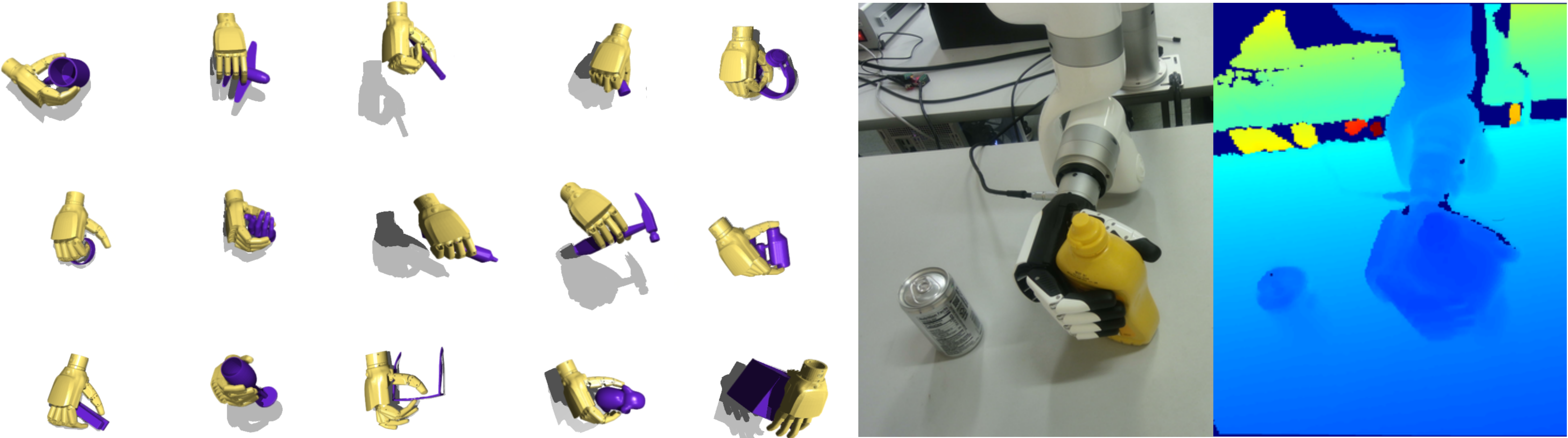}
\caption{We propose \name, a unified control policy that tracks diverse hand-object MoCap references on a dexterous robotic hand~\cite{inspire}. 
\textit{Left}: By using human demonstrations as soft references, we train a robot hand to discover motions that align with its physical form and a given task's intent through large-scale reinforcement learning.
\textit{Right}: We transfer the learned policy into a skill-conditioned generative control policy that takes a partial depth image as input, and successfully deploy the policy on a real robot system.
\label{fig:teaser}}
\end{figure}

\begin{abstract}
    Hand-object motion-capture (MoCap) repositories offer large-scale, contact-rich demonstrations and hold promise for scaling dexterous robotic manipulation. Yet demonstration inaccuracies and embodiment gaps between human and robot hands limit the straightforward use of these data. Existing methods adopt a three-stage workflow, including retargeting, tracking, and residual correction, which often leaves demonstrations underused and compound errors across stages. We introduce \textsc{Dexplore}, a unified single-loop optimization that jointly performs retargeting and tracking to learn robot control policies directly from MoCap at scale. Rather than treating demonstrations as ground truth, we use them as soft guidance. From raw trajectories, we derive adaptive spatial scopes, and train with reinforcement learning to keep the policy in-scope while minimizing control effort and accomplishing the task. This unified formulation preserves demonstration intent, enables robot-specific strategies to emerge, improves robustness to noise, and scales to large demonstration corpora. We distill the scaled tracking policy into a vision-based, skill-conditioned generative controller that encodes diverse manipulation skills in a rich latent representation, supporting generalization across objects and real-world deployment. Taken together, these contributions position \textsc{Dexplore} as a principled bridge that transforms imperfect demonstrations into effective training signals for dexterous manipulation.
\end{abstract}
\input{sec/intro}
\input{sec/relate}
\input{sec/method}

\input{sec/exp}
\input{sec/con}
\input{sec/limitation}

\paragraph{Acknowledgments.} We thank the Toyota Research Institute for the partial support of the robotic hardware used in this research. S. Xu and Y.-X. Wang are supported in part by NSF Grant 2106825 and NIFA Award 2020-67021-32799. This work used computational resources, including the NCSA Delta and DeltaAI and the PTI Jetstream2 supercomputers through allocations CIS230012, CIS230013, and CIS240311 from the Advanced Cyberinfrastructure Coordination Ecosystem: Services \& Support (ACCESS) program, as well as the TACC Frontera supercomputer and Amazon Web Services (AWS) through the National Artificial Intelligence Research Resource (NAIRR) Pilot.

\bibliography{example}

\input{sec/supp}
\end{document}

%% file: sec/intro.tex
\section{Introduction}
Achieving human-level dexterity remains a fundamental and long-standing challenge in robotics and autonomous systems, with broad implications for tasks ranging from object sorting and packaging to food preparation and assisted living \cite{jiang2024dexmimicen, Hao2024SopeDex, bhirangi2023all}. Despite decades of progress, robotic manipulators still struggle with many tasks that humans perform effortlessly \cite{billard2019trends, kroemer2021review}, highlighting the persistent gap between human and robotic manipulation capabilities.

A natural path toward closing this gap is to learn control policies from human demonstrations. However, translating human data into effective robot control is nontrivial: human hands combine high-dimensional kinematics, compliance, and dense tactile sensing; by contrast, robotic hands often differ markedly in morphology, provide fewer independently actuated degrees of freedom, and offer limited sensing and force control. This embodiment mismatch renders direct transfer inherently challenging. To address this, prevailing approaches retarget human demonstrations to the robot’s kinematics, track them with a low-level controller, and add residual or corrective terms to compensate for tracking errors and embodiment gaps~\cite{wang2024dexcap,li2025maniptrans,liu2025dextrack}. However, retargeting errors can propagate and bias downstream learning, for example, a grasp that is feasible for a human may be infeasible or strongly suboptimal for a robotic hand when, if, \textit{e.g.}, strict fingertip correspondence is enforced.

In this work, we propose \textsc{Dexplore}, a paradigm that avoids strict retargeting and post-hoc residual correction. The core idea is to treat demonstrations as soft references that preserve intent while allowing the robot to discover motions compatible with its own embodiment. Concretely, during contact-rich segments, we replace rigid kinematic tracking with adaptive, reference-scoped termination envelopes: at each timestep, the reference induces \textit{spatial scopes} within which a rollout is considered successful. Training begins with wide envelopes and progressively tightens them based on observed success rates, encouraging early exploration and promoting precise tracking whenever feasible. 
To enable real-world deployment under partial observations, we then distill the learned tracker into a vision-based, skill-conditioned generative control policy: latent skill embeddings capture high-level manipulation intent, and a decoder produces low-level actions conditioned on these latent codes. Training combines imitation from a teacher tracker with distribution matching between privileged and partial state encodings. This structure yields a scalable policy that captures diverse manipulation skills (left of Figure~\ref{fig:teaser}) and robustly handles partial observability.

We validate this design in real-world deployment (right of Figure~\ref{fig:teaser}) by running the distilled, vision-based policy on a physical dexterous robotic hand using only single-view depth and proprioception at test time. The policy closes the loop at typical control rates on a standard workstation and issues low-level commands to the hand controller; no mocap references, pose estimators, or force sensors are required at runtime. Conditioned on a compact skill code specifying high-level intent, the policy executes grasping with embodiment-aware adaptation.

Our contributions are threefold: \textbf{(I)} Our \textsc{Dexplore} is a unified single-loop optimization that learns dexterous manipulation directly from human MoCap by treating demonstrations as soft references within adaptive spatial scopes, without explicit retargeting and residual correction. \textbf{(II)} We distill the learned state-based tracker into a vision-based, skill-conditioned generative control policy that maps single-view depth and proprioception, together with a latent skill code, to low-level actions. \textbf{(III)} We demonstrate successful real-world deployment on a dexterous hand using only single-view depth sensing. Overall, \textsc{Dexplore} redefines the role of MoCap data in learning dexterous manipulation, as adaptable guidance that bridges demonstrations and robot-executable skills.

%% file: sec/relate.tex
\section{Related Work}

\textbf{Learning dexterous manipulation from demonstrations.}
Human demonstrations are a powerful substrate for learning dexterous skills, typically collected via motion capture (MoCap) or teleoperation and leveraged through imitation learning (IL). At scale, community platforms and toolkits facilitate collecting and exploiting large demonstration corpora \cite{mandlekar2018roboturk, mandlekar2021robomimic}. Building on this foundation, recent approaches for multi-fingered hands show that closely aligning robot control with demonstrations can reproduce complex hand manipulation behaviors \cite{arunachalam2023dexterous, chen2022dextransfer, zhou2024learning, li2023dexdeform, wang2023mimicplay, an2025dexterous, ye2025dex1b, ye2023learning, qin2022one}. Generative IL has further improved robustness and long-horizon control, via diffusion policies and action-chunking transformers \cite{chi2023diffusionpolicy, zhao2023act} and, more recently, vision–language–action models \cite{luo2025being, yang2025egovla}. However, learning directly from human data, rather than robot demonstrations, still remains challenging because of embodiment mismatches between human and robotic hands.

\textbf{Human-to-robot data transfer.}
To bridge the human-robot embodiment gap, human-to-robot transfer typically proceeds via three routes: (\textbf{I}) video-to-robot pipelines that infer hand motion from human videos and synthesize robot trajectories \cite{qin2022dexmv}; (\textbf{II}) vision-based and AR/VR teleoperation that records robot-compatible demonstrations \cite{handa2020dexpilot, chen2024arcap, cheng2024open, ding2024bunny, zhao2023learning, wu2024gello, he2024learning}; and (\textbf{III}) task-aware retargeting followed by low-level tracking \cite{qin2023anyteleop, wang2024dexcap}. While effective, these routes can be labor-intensive (teleoperation), sensitive to perception noise and occlusion (video/vision-based retargeting), and vulnerable to kinematic and force-control mismatches that accumulate during tracking. In contrast, \textsc{Dexplore} relaxes strict human-to-robot adherence, preserving demonstrator intent while encouraging embodiment-consistent motion discovery.

\textbf{Dexterous manipulation via reinforcement learning.}
Reinforcement learning (RL) provides an alternative to direct imitation, and dexterous skills can be learned even with limited or no demonstration supervision. In practice, demonstrations are often used to bootstrap RL and improve stability and sample efficiency, via on-policy fine-tuning initialized by behavior cloning \cite{rajeswaran2017learning} or off-policy integration using Q-filtering and prioritized replay \cite{nair2018overcoming, vecerik2017ddpgfd}. Sim-to-real systems such as the Rubik’s Cube hand mitigate sensing and dynamics gaps through domain randomization and hindsight experience replay \cite{akkaya2019rubiks, andrychowicz2017her, tobin2017domain}. More recent advances in policy-guided RL from demonstrations and related variants continue to raise in-hand dexterity performance \cite{dasari2023pgdm, wu2023learning, zhu2019dexterous, khandate2023sampling, lum2024dextrah, singh2024dextrah, wan2023unidexgrasp++, xu2023unidexgrasp, zhang2025robustdexgrasp, huang2025fungrasp}. Nevertheless, state-of-the-art pipelines such as DexTrack \cite{liu2025dextrack} and ManipTrans \cite{li2025maniptrans} follow multi-stage cascades that involve explicit human-to-robot transfer, tracking, and residual correction, which can underutilize the demonstration signal and propagate errors across stages. In contrast, our \textsc{Dexplore} integrates reference guidance and reinforcement learning within a single optimization loop, unifying data retargeting, curation, and policy learning.

%% file: sec/method.tex
\section{Methodology}
\begin{figure}
    \centering
    \includegraphics[width=\linewidth]{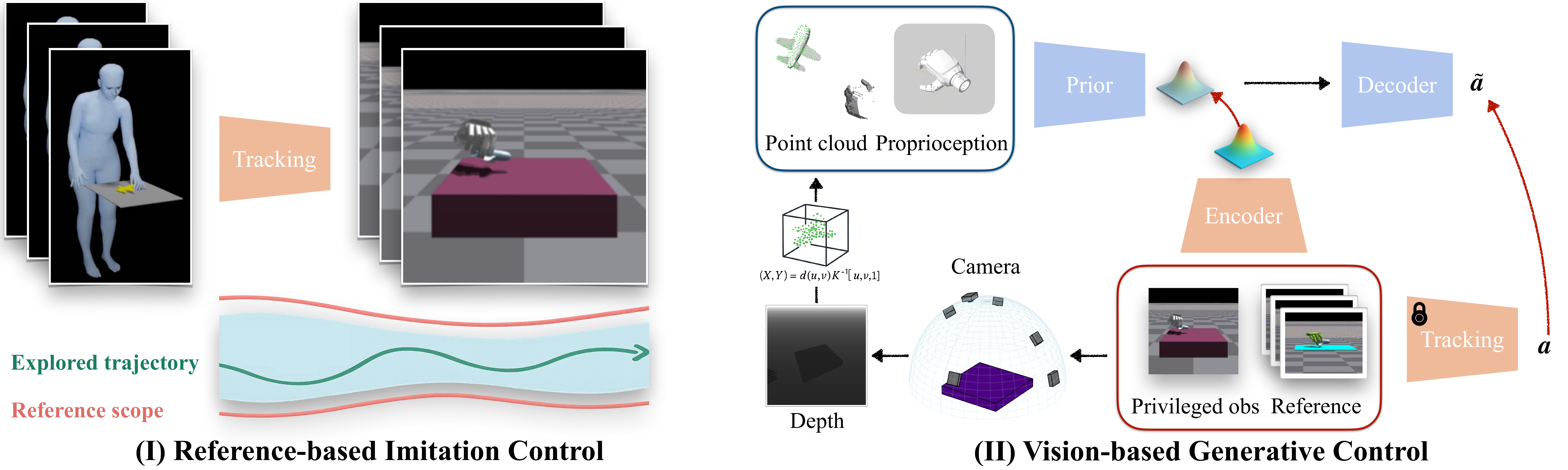}
    \caption{\textbf{Overview of \textsc{Dexplore}.} (\textbf{I}) We first train a state-based policy that acquires scalable dexterous manipulation skills across diverse objects from raw MoCap data, without relying on external retargeting. Rather than rigidly imitating demonstration trajectories, the robot is encouraged to explore within reference-scoped regions, allowing embodiment-specific strategies to emerge.  (\textbf{II}) We then distill these skills into a vision-based policy that embeds diverse manipulation behaviors within a unified latent space. By sampling from this latent representation and integrating perception with hand proprioception, the policy generates human-like behaviors that generalize across tasks.}
    \label{fig:overview}
    \vspace{-1em}
\end{figure}

As illustrated in Figure~\ref{fig:overview}, Dexplore learns dexterous manipulation from human demonstrations in two stages. First, a state-based imitation control policy (Sec.~\ref{sec:31}) is trained with Reference-Scoped Exploration (RSE), enabling the robot to discover embodiment-specific manipulation strategies. This policy is then distilled into a vision-based generative control policy (Sec.~\ref{sec:vision-policy})

\subsection{Reference-based Imitation Control}\label{sec:31}
\noindent\textbf{Task Formulation.} 
The goal is to learn a policy \(\pi\) that enables a dexterous robot hand to manipulate objects by following reference motions derived from human demonstrations. Due to anatomical and actuation mismatches between the robot and the human hand, the policy must compensate for geometric discrepancies rather than replay the reference verbatim. For the robot hand, the wrist is modeled as a floating root joint in world coordinates, with all remaining joints expressed relative to their parents. All manipulated objects are assumed rigid in the simulation, and their configurations are given by full 6-DoF poses (position and orientation), while we show that our robot system is adaptable for non-rigid objects as shown in Figure~\ref{fig:deform}. We formulate reference imitation as a reinforcement learning problem defined over a Markov Decision Process (MDP). The MDP consists of state representations, robot actions, and transition dynamics, together with a reward function designed to encourage faithful yet feasible tracking of the reference motion. We employ Proximal Policy Optimization (PPO)~\cite{schulman2017proximal} to optimize the policy.

\noindent\textbf{State.} At every timestep \(t\), our policy \(\pi\) acts on a hand proprioception and privileged object observation, \(\boldsymbol{x}_t\), combined with a goal-oriented reference component \(\hat{\boldsymbol{x}}_t\). It aggregates hand kinematics, privileged object positioning, and coarse geometry and tactile cues, namely:
\(
\{\{\boldsymbol{R}_t^{\mathrm{h}}, \boldsymbol{J}_t^{\mathrm{h}}, \boldsymbol{\omega}_t^{\mathrm{h}}, \boldsymbol{v}_t^{\mathrm{h}}\}, \{\boldsymbol{R}_t^{\mathrm{o}}, \boldsymbol{p}_t^{\mathrm{o}}, \boldsymbol{\omega}_t^{\mathrm{o}}, \boldsymbol{v}_t^{\mathrm{o}}\}, \{\boldsymbol{D}_t, \boldsymbol{C}_t\}\},
\)
where \(\boldsymbol{R}_t^{\mathrm{h}}, \boldsymbol{J}_t^{\mathrm{h}}\) represent the joint rotations and positions of the robot's hand, respectively; \(\boldsymbol{\omega}_t^{\mathrm{h}}, \boldsymbol{v}_t^{\mathrm{h}}\) denote the angular and linear velocities of the hand joints. Similarly, \(\boldsymbol{R}_t^{\mathrm{o}}, \boldsymbol{p}_t^{\mathrm{o}}\) represent the object's orientation and position, with \(\boldsymbol{\omega}_t^{\mathrm{o}}, \boldsymbol{v}_t^{\mathrm{o}}\) capturing its angular and linear velocities.
We integrate coarse object geometry cues and simplified contact information through two sensory modalities following~\cite{xu2025intermimic}: (i) \(\boldsymbol{D}_t\), consisting of vectors~\cite{christen2022d} pointing from each hand joint to the nearest surface points of the object, and (ii) \(\boldsymbol{C}_t\), binary contact indicators that reflect whether each rigid body on hand is in contact, mimicking tactile sensing~\cite{dahiya2009tactile}.
The goal reference component \(\hat{\boldsymbol{x}}_t\) is derived from the reference MANO demonstrations, structured as:
\(
\{\hat{\boldsymbol{x}}_{t+k}\}_{k \in K},
\)
with each future state \(\hat{\boldsymbol{x}}_{t+k}\) defined explicitly as:
\(
\{\{\boldsymbol{M}(\hat{\boldsymbol{R}}_{t+k}^{\mathrm{h}}) \ominus \boldsymbol{M}({\boldsymbol{R}}_t^{\mathrm{h}}), \boldsymbol{M}(\hat{\boldsymbol{J}}_{t+k}^{\mathrm{h}}) - \boldsymbol{M}({\boldsymbol{J}}_t^{\mathrm{h}})\}, \{\hat{\boldsymbol{R}}_{t+k}^{\mathrm{o}} \ominus \boldsymbol{R}_t^{\mathrm{o}}, \hat{\boldsymbol{p}}_{t+k}^{\mathrm{o}} - {\boldsymbol{p}}_t^{\mathrm{o}}\},\{\boldsymbol{M}(\hat{\boldsymbol{D}}_{t+k}) - \boldsymbol{M}({\boldsymbol{D}}_t), \boldsymbol{M}(\hat{\boldsymbol{C}}_{t+k}) - \boldsymbol{M}({\boldsymbol{C}}_t\}), \{\hat{\boldsymbol{R}}_{t+k}^{\mathrm{h}}, \hat{\boldsymbol{J}}_{t+k}^{\mathrm{h}}, \hat{\boldsymbol{R}}_{t+k}^{\mathrm{o}}, \hat{\boldsymbol{p}}_{t+k}^{\mathrm{o}}\}\},
\)
with \(\ominus\) indicating a rotation difference. Here, \(K\) specifies the reference indices that define the goal horizon. Because the robot hand and the demonstration state \((\hat{\cdot})\) are defined on different skeletal structures, we introduce a mapping \(\boldsymbol{M}\) that projects both representations onto a common subset of key joints before computing deltas. This mapping plays a role analogous to the correspondence used in retargeting objectives, and we adopt the one provided in~\cite{qin2023anyteleop}. For the Inspire hand~\cite{inspire}, for example, the mapping simplifies to the five fingertips. All continuous components are expressed in a coordinate frame anchored at the hand’s floating root joint and canonicalized by its current rotation.

\noindent\textbf{Action.} The control is divided into two parts: a 6-DoF floating-wrist signal and a set of local finger commands. 
Each element is a proportional-derivative (PD) target in which rotations are expressed with an exponential-map parametrization. These targets are converted to joint torques to drive the robot.  
For kinematically coupled (mimic) joints, \textit{e.g.}, in the Inspire hand, the controller scales the driving joint’s PD target by its predefined coupling coefficient (originally specified for joint positions) to reproduce the mechanical linkage in hardware. While this approximation may not perfectly capture the true coupling dynamics, it provides a practical control strategy. 
The policy operates in a residual action space for floating-wrist control, where a 6-DoF positional offset is added to the current wrist state to form the PD target, preventing the network from having to learn unbounded absolute trajectories and thereby improving stability and generalization.

\noindent\textbf{Reward.}  
At each timestep, the reward consists of two complementary components: a state-reference matching term and an energy regularization term. The matching term, \(R_{\text{match}}(\boldsymbol{x}_{t}, \hat{\boldsymbol{x}}_{t})\), encourages the simulated state \(\boldsymbol{x}_{t}\) to align with the demonstration \(\hat{\boldsymbol{x}}_{t}\). Following~\cite{xu2025intermimic}, \(R_{\text{match}}\) combines (i) kinematic alignment of hand joint rotations and positions \((R_{\boldsymbol{R}}^{\mathrm{h}}, R_{\boldsymbol{J}}^{\mathrm{h}})\), and object orientation and position \((R_{\boldsymbol{R}}^{\mathrm{o}}, R_{\boldsymbol{p}}^{\mathrm{o}})\); and (ii) coarse geometric \(R_{\boldsymbol{D}}\) and contact \(R_{\boldsymbol{C}}\) correspondences. To discourage excessive actuation, we add energy terms \(R_{\text{energy}}(\boldsymbol{a}_{t})\), including a penalty for large changes in consecutive PD targets as well as high joint accelerations and velocities, thereby promoting smooth and energy-efficient motion.  
The reward formulation below is designed to support retargeting by emphasizing correspondence in behaviorally relevant features, enabling robust transfer across embodiments.

\noindent\textit{Kinematic State Matching.}  
The primary objective is to align the robot hand with demonstration, emphasizing the matching of joints and links specified by the mapping \(\boldsymbol{M}\), as discussed in state formulation. Formally, we define
\(
R_{\boldsymbol{J}}^{\mathrm{h}} = \exp\!\Big(- \sum \lambda_{\boldsymbol{J}}^{\mathrm{h}} \big\| \boldsymbol{M}(\hat{\boldsymbol{p}}) - \boldsymbol{M}(\boldsymbol{p}) \big\|^2 \Big),
R_{\boldsymbol{R}}^{\mathrm{h}} = \exp\!\Big(- \sum \lambda_{\boldsymbol{R}}^{\mathrm{h}} \big\| \boldsymbol{M}(\hat{\boldsymbol{R}}) \ominus \boldsymbol{M}(\boldsymbol{R}) \big\|^2 \Big),
\)
where $\lambda_{\boldsymbol{J}}^{\mathrm{h}}, \lambda_{\boldsymbol{R}}^{\mathrm{h}}$ are weighting hyperparameters.

\noindent\textit{Dynamic State Matching.}  
The policy is also encouraged to maintain appropriate interaction dynamics, such as keeping fingertips near functional affordances and preserving contact correspondences. The associated rewards are
\(
R_{\boldsymbol{D}} = \exp\!\big(-\lambda_{\boldsymbol{D}} \|\hat{\boldsymbol{D}} - \boldsymbol{D}\|^2\big), 
R_{\boldsymbol{C}} = \exp\!\big(-\lambda_{\boldsymbol{C}}\|\hat{\boldsymbol{C}} - \boldsymbol{C}\|^2\big)
\). These terms promote consistent manipulation with embodiment gaps preventing exact kinematics mimic.

\noindent\textbf{Early Termination.}
We employ early termination criteria to halt episodes when simulation significantly deviates from feasible or desired trajectories. Formally, termination conditions are defined as:
\(
\text{(\textbf{I})} R_{\boldsymbol{J}}^{\mathrm{h}} < \kappa_{\boldsymbol{J}}^{\mathrm{h}},\text{(\textbf{II})} R_{\boldsymbol{p}}^{\mathrm{o}} < \kappa_{\boldsymbol{p}}^{\mathrm{o}},\text{(\textbf{III})} R_{\boldsymbol{R}}^{\mathrm{o}} < \kappa_{\boldsymbol{R}}^{\mathrm{o}},\text{(\textbf{IV})}R_{\boldsymbol{D}} < \kappa_{\boldsymbol{D}}, \text{(\textbf{V})} \boldsymbol{C}_t \ne \hat{\boldsymbol{C}}_t,\forall t \in [t_0, t_0+10],
\)
These criteria promptly terminate episodes when interactions deviate irrecoverably from reference trajectories, thus facilitating effective policy learning via collecting meaningful experiences~\cite{peng2018deepmimic}.

\noindent\textbf{State Initialization.}  
Because of the anatomical and kinematic differences between the robot and the MANO hand model~\cite{MANO}, directly assigning MANO poses as initial robot states is infeasible. At the start of training, we initialize only the global rotation and translation of the robot palm according to the reference pose, while setting all other degrees of freedom to zero. Following~\cite{xu2025intermimic}, we then cache high-quality rollout states and subsequently reuse them as initialization. To further improve training efficiency, we adopt prioritized sampling: rollouts are more likely to start from challenging phases rather than motions for hand approaching.

\noindent\textbf{Learning with Reference Scope.}  
Our key idea is to enable flexible trajectory exploration within a unified imitation learning framework that integrates retargeting, tracking, and correction. The central mechanism is to reduce reliance on strict tracking rewards and instead shape the policy through adaptive early termination criteria.  
(\textbf{I}) We introduce a reward weighting scheme in which the kinematic matching weights \(\lambda_{\boldsymbol{R}}^{\mathrm{h}}\) and \(\lambda_{\boldsymbol{J}}^{\mathrm{h}}\) are proportional to the hand–object surface distance, while the energy penalty \(R_{\text{energy}}(\boldsymbol{a}_{t})\) is inversely proportional to this distance. This design encourages smoother, lower-energy motions when tracking constraints are relaxed. For example, the kinematic weight is given by  
\(
w(\boldsymbol{D}) = \min\!\Bigl(1, \tfrac{\boldsymbol{D}}{0.20\,\text{m}}\Bigr).
\)
Although these weights vary across timesteps, they are deterministic functions of the reference trajectory and therefore the reward still remains stationary from the policy’s perspective, ensuring that the optimization target is consistent, and thus, the varying weights do not bias the policy toward unintended behaviors.
(\textbf{II}) We shape the reward via adaptive early termination thresholds. For each termination criterion \(\kappa\), we initialize with a large value and progressively tighten it according to the ratio of failed rollouts:  
\(
\kappa = \kappa^{\text{init}} \cdot \frac{N_{\text{fail}}}{N_{\text{total}}},
\)  
where \(\kappa^{\text{init}}\) is the initial large threshold, \(N_{\text{fail}}\) is the number of unsuccessful rollouts (early termination events), and \(N_{\text{total}}\) is the total number of rollouts visited at the current frame. 
\begin{figure}
    \centering
    \includegraphics[width=\linewidth]{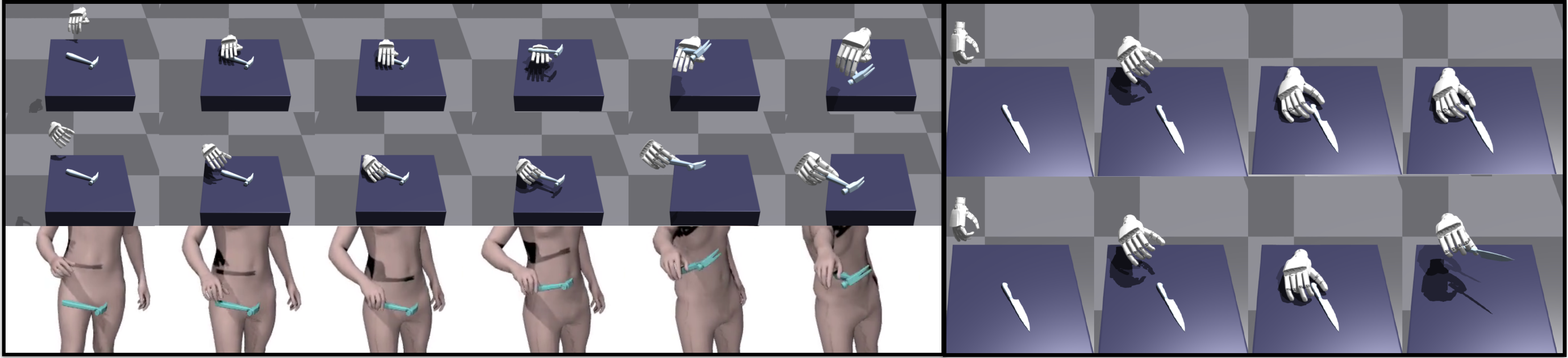}
    \caption{\textit{Left}: Comparison with AnyTeleop~\cite{qin2023anyteleop} (\textbf{top}) and retargeted MoCap data (\textbf{bottom}), both without task awareness. For hands with limited DoFs, retargeting often yields unnatural poses, unlike our tracking results (\textbf{middle}). \textit{Right}: Comparison with DexTrack~\cite{liu2025dextrack} (\textbf{top}). For objects with small grasp regions, the baseline fails to grasp reliably, while our method (\textbf{bottom}) succeeds.}

    \label{fig:baseline}
    \vspace{-1.5em}
\end{figure}
\subsection{Vision-based Generative Control}\label{sec:vision-policy}
\noindent\textbf{Task Formulation.}
The objective of generative control is to learn a policy $\tilde\pi$ that enables a robotic hand to manipulate objects using partial observations and optionally sparse goals. 

\noindent\textbf{Observation.} As illustrated in Figure~\ref{fig:overview} II), the partial observations consist of (i) hand proprioception ($\tilde{\boldsymbol{x}}$), including the global wrist position and orientation as well as local joint rotations, and (ii) an object point cloud ($\boldsymbol{P}$) reconstructed from a single-view depth image. Depth images are captured by a camera randomly positioned on a hemisphere above the table, with intrinsic parameters matched to our Femto Bolt sensor~\cite{femtobolt}. Each depth map is converted into a 3D point cloud using the known camera model.  
To encode the visual input, we employ a PointNet++~\cite{qi2017pointnet++} backbone that jointly processes the object point cloud together with observed human joint positions from the camera, yielding a compact representation of both object geometry and hand-object interaction context. Other proprioceptive signals, such as joint velocities and tactile information, are excluded due to their large sim-to-real gap. In addition, the observation includes short-horizon historical states and a binary indicator specifying whether the current timestep occurs before or after the onset of manipulation.  

\noindent\textbf{Goal.} Following~\cite{tessler2024maskedmimic}, we define sparse goals by selectively unmasking portions of the reference motion, while the remaining components are either masked during training or replaced by mask tokens at inference. For example, unmasking a desired wrist trajectory, obtained from reference data during training or generated by an RRT-based motion planner in real-world evaluation, serves as the goal condition. The policy then leverages its latent representation to infer the masked components and generate diverse, human-like manipulation behaviors.

\input{tab/quantitative}

\noindent\textbf{Learning the Generative Control.}  
Introducing a latent embedding is essential in order to (i) capture and modulate diverse behaviors, (ii) eliminate reliance on complete reference trajectories, and (iii) model the uncertainty that arises from missing privileged information at inference time. Our framework consists of an encoder \( q_\phi(z \mid \boldsymbol{x}, \hat{\boldsymbol{x}}) \), a prior network \( p_\psi(z \mid \tilde{\boldsymbol{x}}, \boldsymbol{P}) \), and a decoder policy \( \tilde\pi_\theta(\tilde{\boldsymbol{a}} \mid z, \tilde{\boldsymbol{x}}) \).  
The encoder takes both the privileged robot-object state \(\boldsymbol{x}\) and reference \(\hat{\boldsymbol{x}}\), and outputs a Gaussian latent distribution \(\mathcal{N}(\boldsymbol{\mu}_q,\boldsymbol{\Sigma}_q)\). In parallel, the prior network produces a \emph{state-dependent} Gaussian distribution \(\mathcal{N}(\boldsymbol{\mu}_p, \boldsymbol{\Sigma}_p)\).
At each timestep, the latent skill embedding is sampled as  
\(
z = \boldsymbol{\mu}_p + \boldsymbol{\mu}_q + \boldsymbol{\Sigma}_q^{1/2}\epsilon, \epsilon \sim \mathcal{N}(0, I),
\)
with the noise \(\epsilon\) held fixed within an episode to ensure temporal consistency of the executed skill, following~\cite{tessler2024maskedmimic,yao2022controlvae}. 
The framework is trained via online imitation learning using DAgger, with the imitation control policy introduced in Sec.~\ref{sec:31} serving as the expert teacher. It minimizes the reconstruction loss  
\(
\mathcal{L}_\mathrm{rec} = \|\boldsymbol{a} - \tilde{\boldsymbol{a}}\|^2,
\)  
where \(\tilde{\boldsymbol{a}}\) are predicted actions and \(\boldsymbol{a}\) are expert-provided reference actions. A KL regularization term further encourages the encoder distribution to remain close to the prior:  
\(
\mathcal{L}_\mathrm{KL} = D_\mathrm{KL}\!\bigl(\mathcal{N}(\boldsymbol{\mu}_p + \boldsymbol{\mu}_q, \boldsymbol{\Sigma}_q)\;\|\;\mathcal{N}(\boldsymbol{\mu}_p, \boldsymbol{\Sigma}_p)\bigr).
\), ensuring that the encoder does not capture information unavailable at inference.
The overall training objective is  
\(
\mathcal{L} = \mathcal{L}_\mathrm{rec} + \beta\,\mathcal{L}_\mathrm{KL},
\)
with \(\beta\) gradually increased during training to promote a more structured latent space, following~\cite{higgins2017beta}. At inference time, the encoder is omitted, and latents are sampled directly from the learned prior:  
\(
z \sim \mathcal{N}(\boldsymbol{\mu}_p, \boldsymbol{\Sigma}_p),
\)
yielding a generative policy \(\tilde{\pi}\) that produces goal-conditioned dexterous manipulation behaviors using only partial observations.

%% file: tab/quantitative.tex
\begin{table*}
    \centering
    \resizebox{1.0\textwidth}{!}{%
\begin{tabular}{@{\;}llcccc@{\;}}
        \toprule
        Embodiment & Tracking w/ Retargeting (optional) & \makecell[c]{$R_{\text{err}}$ ($\text{rad}, \downarrow$)} & \makecell[c]{$T_{\text{err}}$ (${cm}, \downarrow$)}   & $E_{\text{finger}}$ ($\text{cm}, \downarrow$) & Success Rate ($\%, \uparrow$) \\

        \cmidrule(l{0pt}r{1pt}){1-2}
        \cmidrule(l{2pt}r{2pt}){3-6}

        \multirow{4}{*}{ Inspire~\cite{inspire} } & DexTrack~\cite{liu2025dextrack} w/ AnyTeleop~\cite{qin2023anyteleop} & 0.475 / 0.216 & 4.93 / 2.29  & 6.71 / 5.67  & 7.4
        \\ 
        ~ & DexTrack~\cite{liu2025dextrack} w/ \textbf{Ours} & 0.515 / 0.467 & \textbf{3.75} / 3.26  & 6.43 / 6.28 & 69.8
        \\ 
        
        \cmidrule(l{0pt}r{1pt}){2-2}
        \cmidrule(l{2pt}r{2pt}){3-6}
        
        ~ & \textsc{Dexplore}~(w/o RSE)  & {0.514 / 0.333} & {3.88 / 2.59} & {6.12 / 5.71} & {29.9} %
        \\ 
        ~ & \textsc{Dexplore}  & {\textbf{0.474} / 0.452} & {3.77 / 3.65} & {\textbf{6.06} / 6.04} & {\textbf{87.7}}
         \\ 
        
        \cmidrule(l{0pt}r{1pt}){1-2}
        \cmidrule(l{2pt}r{2pt}){3-6}

         \multirow{4}{*}{Allegro~\cite{allegro} } & DexTrack~\cite{liu2025dextrack} w/ AnyTeleop~\cite{qin2023anyteleop} & 0.595 / 0.471 & 4.21 / 3.39  & 9.38 / 9.36 & 45.9
        \\ 
        ~ & DexTrack~\cite{liu2025dextrack} w/ \textbf{Ours} & 0.560 / 0.512 & 4.79 / 4.32  & 8.00 / 7.91 & 77.4
        \\ 
        
        \cmidrule(l{0pt}r{1pt}){2-2}
        \cmidrule(l{2pt}r{2pt}){3-6}

         ~ & \textsc{Dexplore}~(w/o RSE)  & {0.549 / 0.481} & {\textbf{4.19} / 3.68} & {\textbf{7.98} / 7.92} & {29.9} %
        \\ 
        ~ & \textsc{Dexplore}  & {\textbf{0.494} / 0.454} & {4.48 / 4.06} & {8.10 / 8.01} & {\textbf{78.7}}
        \\ 

        \bottomrule
 
    \end{tabular}
    }
    \caption{ 
    {\textbf{Quantitative evaluations}} comparing our method with the baseline on the GRAB~\cite{taheri2020grab} dataset. Tracking errors are reported as averages computed either over successful rollouts / over all frames. Note that averaging errors across all frames generally yields lower values when the success rate is low, since failed sequences mainly involve easy approaching without manipulating the object.} 
    \label{tb_exp_main}
    \vspace{-2em}
\end{table*}

%% file: sec/exp.tex
\section{Experiments}
We evaluate the state-based policy's ability to imitate human demonstrations and generalize to unseen scenarios. We then assess the vision-based policy framework's effectiveness in achieving manipulation within simulation and its successful transfer to real-world deployment.

\noindent\textbf{Data and Robot Models.} We evaluate our state-based policy using two robotic hands in simulation: (\textbf{I}) the Inspire hand~\cite{inspire}, which is less dexterous than a human hand, featuring 12 degrees of freedom (DoFs) with only 6 actuated. We simulate mimic joint effects in IsaacGym~\cite{makoviychuk2021isaac}, as described in Sec.~\ref{sec:31}; and (\textbf{II}) the Allegro hand~\cite{allegro}, a fully actuated robotic hand with 16 DoFs, notable for its larger size compared to a human hand. These two hands comprehensively test our algorithm's robustness across embodiment gaps in both actuation complexity and physical dimensions.  
We primarily utilize the GRAB~\cite{taheri2020grab} dataset as our source for human demonstrations, selecting 658 sequences involving 51 objects from an original set of 1,269 sequences, ensuring that selected sequences focus on single-hand (primarily right-hand) manipulations without bimanual interactions. Our framework does not require explicit retargeting; however, we evaluate retargeting algorithms separately for comparison purposes. We further investigate the generalizability of our trained policy using novel data from the TACO dataset~\cite{liu2024taco}, evaluating whether our model benefits from scalability and effectively adapts to diverse object geometries.

\begin{figure}
    \centering
    \includegraphics[width=\linewidth]{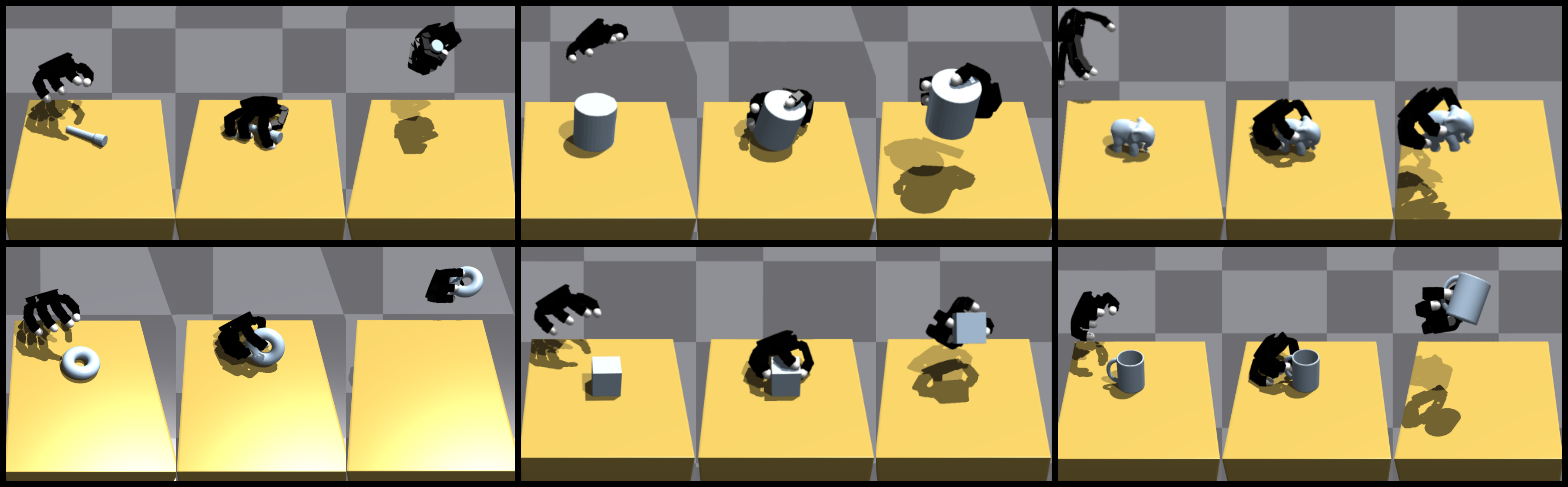}
    \caption{Our imitation control framework is also suitable for Allegro hands~\cite{allegro}. Compared to Inspire hands~\cite{inspire}, Allegro hands feature a different morphology with four fingers instead of five, and offer more degrees of freedom (DoFs), but are significantly larger in size.}
    \label{fig:allegro}
\end{figure}

\begin{figure}
    \centering
    \includegraphics[width=\linewidth]{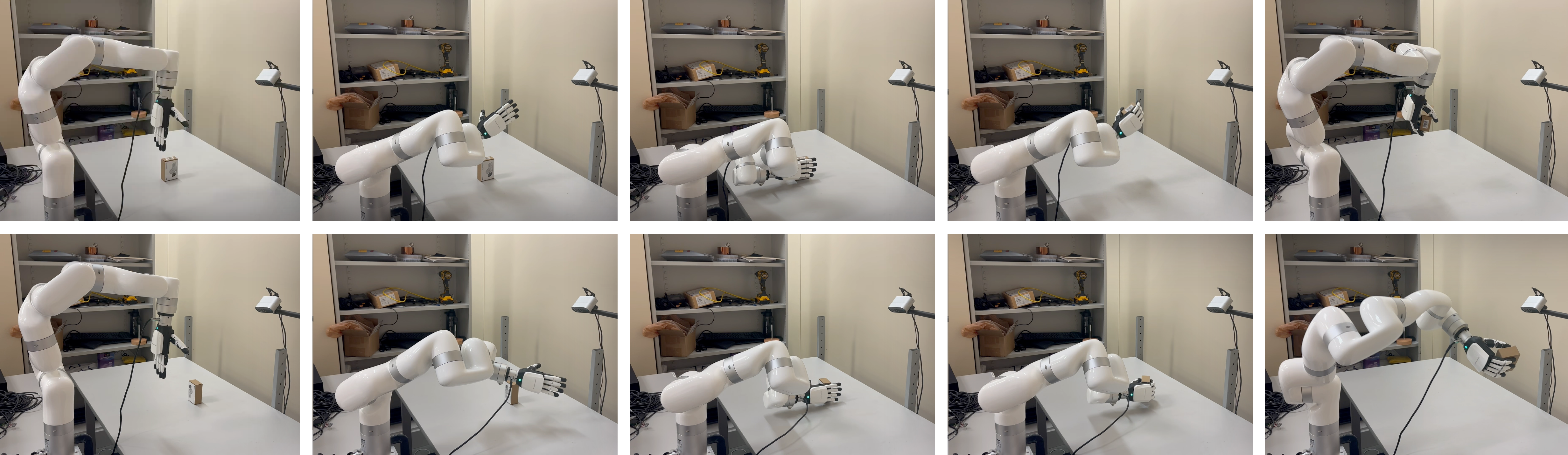}
    \caption{The vision-based policy deployed on an XArm-7 robotic arm~\cite{xarm7} equipped with an Inspire dexterous hand~\cite{inspire} and a Fremto Bolt depth camera~\cite{femtobolt}.}
    \label{fig:real}
\end{figure}
\noindent\textbf{Metrics.} For evaluating the state-based policy, we introduce four metrics to assess tracking accuracy and task success. The \textit{Object Rotation Error} ($R_{\text{err}}$) measures how closely the tracked object's orientation aligns with the reference orientation per frame. The \textit{Object Translation Error} ($T_{\text{err}}$) captures the positional discrepancy between the tracked and reference object positions on a per-frame basis. Additionally, the \textit{Fingertip Error} ($E_{\text{finger}}$) quantifies the average positional difference of fingertips between tracked and reference poses at each frame. Finally, the \textit{Success Rate} is defined as the proportion of tracking attempts completed without dropping the object or exhibiting large deviations as indicated by the aforementioned metrics. To evaluate the vision-based policy, since explicit reference trajectories are unavailable, we rely on the \textit{Success Rate} and \textit{Contact Ratio} metrics. Here, success indicates maintaining stable hand-object contact over a duration and returning the object to the table, while the contact ratio measures the proportion of time the hand remains contact.

\noindent\textbf{Baselines.}
We consider two types of baselines:
(\textbf{I}) AnyTeleop~\cite{qin2023anyteleop}: a retargeting algorithm for robot hands from human demonstrations. We compare our policy's simultaneous retargeting and tracking capabilities to the kinematic replay from AnyTeleop's retargeting.
(\textbf{II}) DexTrack~\cite{liu2025dextrack}: We adapt it for comparison with our imitation control policy. Since DexTrack's primary contributions, including homotopy optimization and the combination of imitation learning (IL) and reinforcement learning (RL), have not been publicly released, we adapt the available components and train them in alignment with our experimental setting. Given DexTrack requires pre-processed retargeting, we also provide the high-quality imitation results from our \textsc{Dexplore} as the source of their training. 

\noindent\textbf{Quantitative Evaluation.} 
Table~\ref{tb_exp_main} shows that our method consistently outperforms baseline methods on each robot hand. Our \textsc{Dexplore} achieves notably higher success rates. The ablation study on RSE demonstrates the effectiveness of scope exploration in compensating for robot-specific strategies. Additionally, since our policy is designed with retargeting in mind, we observe substantial performance improvements in the baseline DexTrack when utilizing rollouts from our policy as a reference. 
Table~\ref{tb:ablation} evaluates our vision-based policy. Even with randomly sampled camera views, which might lead to severe hand-object occlusion, without privileged information, the policy still demonstrates strong performance in object manipulation and maintaining contact. 

\noindent\textbf{Qualitative Evaluation.}
We compare our \textsc{Dexplore} against the DexTrack baseline~\cite{liu2025dextrack} in Figure~\ref{fig:baseline}, with the example where the affordance region is particularly constrained, requiring delicate and precise movements to grasp a knife. Our approach successfully accomplishes this challenging task. In Figure~\ref{fig:baseline}, we also highlight that task-unaware retargeting~\cite{qin2023anyteleop}, which optimize fingertip distances as key features, is highly sensitive when the robot hand has fewer DoFs compared to a human hand, since certain finger flexions easily achievable by humans are infeasible for robot hands, resulting in unrecoverable retargeting failures. In contrast, our \textsc{Dexplore} effectively discovers solutions that achieve natural manipulation motions, albeit slightly deviating from the original trajectories. Furthermore, our algorithm remains effective even with increased DoFs, as demonstrated by the Allegro hand example in Figure~\ref{fig:allegro}. Our imitation control policy, enhanced by inherently encoded geometric information and guided exploration allowing deviation from the reference, demonstrates strong generalization capabilities to settings unseen during training -- such as manipulating larger and heavier objects, as illustrated in Figure~\ref{fig:large}.
For the distilled vision-based generative control, we evaluate it under conditions where object geometries are entirely unseen during training. As demonstrated in Figure~\ref{fig:taco}, our policy successfully generalizes to manipulate these novel geometries. Additionally, skills from human demonstrations are distilled into a rich latent space learned from the MoCap data. By sampling from this latent space, the policy generates diverse and plausible manipulation styles, as illustrated in Figure~\ref{fig:diverse}. We further validate this policy for real-world deployment, with successful examples shown in Figures~\ref{fig:real} and~\ref{fig:deform}.

%% file: sec/con.tex
\section{Conclusion}
We introduce \textsc{Dexplore}, a novel paradigm for scalable dexterous robotic manipulation that unifies retargeting and tracking through learning from human demonstrations. Rather than rigidly replicating trajectories, our approach leverages flexible guidance, enabling robots to discover natural and efficient manipulation strategies adapted to their morphology. The learned state-based imitation policy is distilled into a vision-based generative control policy, removing reliance on privileged object observations and dense human references, thereby facilitating sim-to-real transfer. Experiments show that our policy outperforms baselines and that the distilled visual policies generalize effectively and are deployable in real-world scenarios.

%% file: sec/limitation.tex
\section{Limitation}
Our approach currently faces several limitations. First, it sometimes struggles with reliably manipulating very small or thin objects, where slight inaccuracies in fingertip positioning significantly affect success. Second, we focus exclusively on single-hand interactions; while extension to multi-hand manipulation is straightforward, coordinated behaviors require further exploration. Additionally, although our method directly learns from MoCap data for scalability, our training dataset is relatively modest, limiting generalization. Future work includes training on larger, more diverse datasets, and extending our approach to coordinated multi-hand manipulation scenarios.

%% file: sec/supp.tex
\clearpage
\setcounter{page}{1}
\vbox{%
  \hsize\textwidth
  \linewidth\hsize
  \vskip 0.1in
  \centering
  {\LARGE\bf \name: Scalable Neural Control for Dexterous Manipulation from Reference‑Scoped Exploration \par}
  \vskip 0.2in
  {\Large \bf Appendix\par}
  \vspace{0.3in minus 0.1in}
}

\vspace{2em}

\setcounter{table}{0}
\renewcommand{\thetable}{\Alph{table}}
\renewcommand*{\theHtable}{\thetable}
\setcounter{figure}{0}
\renewcommand{\thefigure}{\Alph{figure}}
\renewcommand*{\theHfigure}{\thefigure}
\setcounter{section}{0}
\renewcommand{\thesection}{\Alph{section}}
\renewcommand*{\theHsection}{\thesection}

\noindent In this appendix, we provide additional experimental setups:
\begin{enumerate}
    \item \textbf{Demo Video.} A demonstration video including the real world experiment is provided, as described in Sec.~\ref{sec:demo}.
    \item \textbf{Simulation Setup.} The environment configuration for simulation is introduced in Sec.~\ref{sec:phys_para}.
    \item \textbf{Real-world Setup.} Sec.~\ref{sec:add_exp} presents further details on our real-world deployment.
\end{enumerate}

\section{Demo Video} \label{sec:demo}

In addition to the qualitative results presented in the main paper, we provide a demo video (\url{https://sirui-xu.github.io/dexplore/demo.mp4}) for more detailed visualizations of the tasks, further illustrating the efficacy of our approach. The demo video conveys the following key points:

\begin{enumerate}
\item We demonstrate that our vision-based policy successfully transfers to real-world dexterous grasping tasks.
\item Our state-based policy consistently outperforms baseline methods, such as DexTrack~\cite{liu2025dextrack} and AnyTeleop~\cite{qin2023anyteleop}, demonstrating superior task execution and generalization capabilities, such as to novel objects.
\item Our holistic method integrates retargeting and skill learning into a unified, cohesive framework, enhancing overall adaptability and performance.
\item Our framework exhibits significant versatility, successfully adapting to different robotic embodiments, including the Allegro~\cite{allegro} hand larger than hand hands, and Inspire~\cite{inspire} hand with unactuated joints.
\item Our vision-based policy, which operates without privileged information, reliably accomplishes tasks with high robustness, emphasizing the practicality and resilience of our approach in real-world scenarios.
\end{enumerate}

\section{Simulation Setup} \label{sec:phys_para}
We use MANO models~\cite{MANO} to represent human hand reference data, and objects are represented using convex decomposition into $20$ convex hulls for efficient simulation. Key simulation parameters are summarized concisely in Table~\ref{tab:physics_hyper}.

\begin{figure}[h]
    \centering
\includegraphics[width=\textwidth]{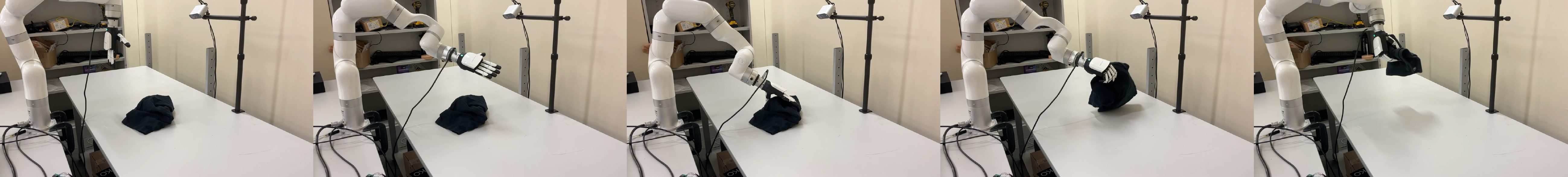}
\captionof{figure}{our framework can successfully grasp deformable objects (\textit{i.e.}, cloth).
\label{fig:deform}}

\end{figure}
\begin{figure}
    \centering
    \includegraphics[width=\linewidth]{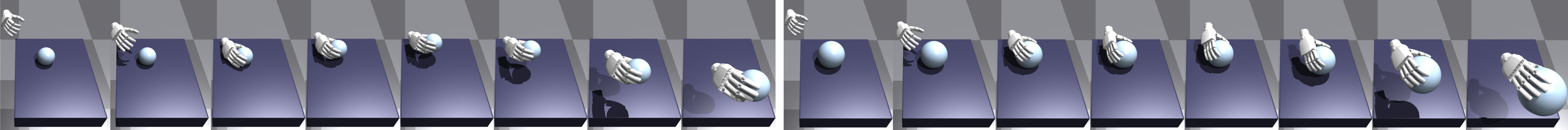}
    \caption{Our imitation control policy generalizes effectively to manipulating larger and heavier objects unseen during training. Specifically, we scale the object's size by a factor of 1.5, resulting in a corresponding weight increase of $1.5^3$ shown on the right.}
    \label{fig:large}
\end{figure}
\begin{figure}
    \centering
    \includegraphics[width=\textwidth]{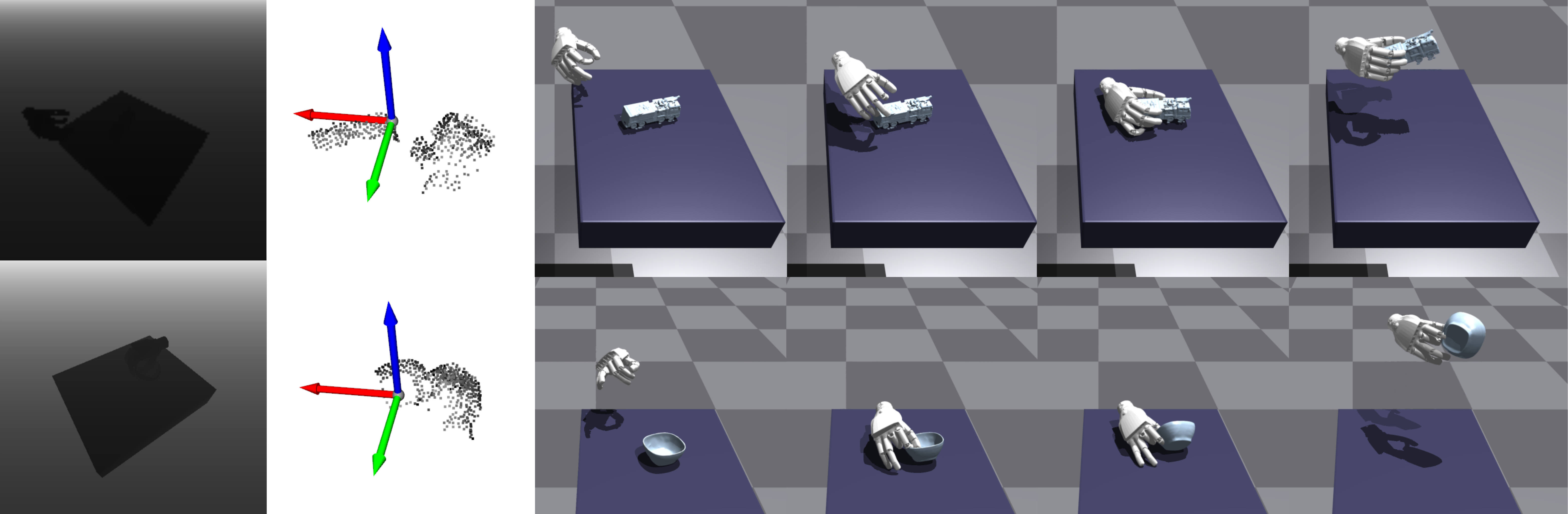}
    \caption{Our vision-based generative control, though trained exclusively on the GRAB dataset~\cite{taheri2020grab}, generalizes to manipulating novel objects, specifically, object 081 and object 194 from the TACO dataset~\cite{liu2024taco}.}
    \label{fig:taco}
\end{figure}

\begin{table}
\centering
\small
\begin{tabular}{ll}
\toprule
Parameter & Value \\[0.5ex]
\midrule
Sim timestep & $1/60$ s \\[0.2ex]
Control timestep & $1/30$ s \\[0.2ex]
Environment count & 4096 \\[0.2ex]
Physics substeps & 8 \\[0.2ex]
Position solver iterations & 8 \\[0.2ex]
Velocity solver iterations & 1 \\[0.2ex]
Contact offset & 0.02 \\[0.2ex]
Rest offset & 0.0 \\[0.2ex]
Max depenetration vel. & 10 \\[0.2ex]
Object restitution & 0.7 \\[0.2ex]
Object friction & 0.9 \\[0.2ex]
Object density & 50 \\[0.2ex]
\bottomrule
\end{tabular}
\vspace{0.5em}
\caption{Summary of simulation parameters in Isaac Gym~\cite{makoviychuk2021isaac}.}
\label{tab:physics_hyper}
\end{table}

\section{Real-World Deployment} \label{sec:add_exp}
\noindent\textbf{Hardware Setup.} As illustrated in Figure~\ref{fig:teaser} of the main paper, our experimental setup includes an Xarm-7 robotic arm~\cite{xarm7} coupled with an Inspire RH56DFTP dexterous right hand~\cite{inspire}. The Xarm-7 has 7 independently motor-driven joints, while the Inspire RH56DFTP hand comprises 12 joints actuated by 6 motors, meaning 6 joints serve as mimic joints. In simulation, the proportional-derivative (PD) targets for these mimic joints are proportionally scaled based on the targets of their actuated counterparts, as detailed in Sec.~\ref{sec:31} of the main paper. In our real-world experiments, we directly utilize the PD targets from the actuated joints for robot control. The real robot employs internally defined position control parameters for proportional gain and derivative gain, whereas in simulation, we explicitly set these values to 100 and 10, respectively. Additionally, we utilize a Femto Bolt depth camera~\cite{femtobolt} to perform depth estimation and extract point cloud data.

\begin{figure}
    \centering
    \includegraphics[width=\textwidth]{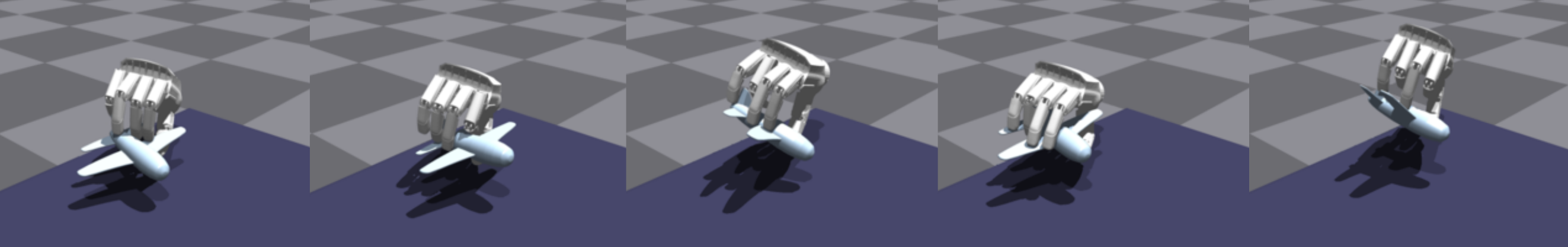}
    \caption{Our vision-based generative control formulates a latent space for sampling, resulting in diverse manipulation styles. We visualize poses at the moment when the hand makes initial contact with the object.}
    \label{fig:diverse}
\end{figure}

\noindent\textbf{Deployment.} To achieve smooth and efficient robot motion during deployment, we first apply forward kinematics to calculate the robot arm point cloud. We then filter the resulting scene to retain only the hand and object point cloud, removing background, table, and arm points. During execution, and after the hand enters the camera’s view, there can be a noticeable gap between the robot’s proprioceptive estimate of hand joint global positions and the observed point cloud. To mitigate this, we map each joint to the nearest points in the point cloud as its representation, reducing the discrepancy and making the policy more aware of the robot–object relative positions.

We do not directly apply every step of the 30 fps rollout in the approach and lifting stages. Instead, the full rollout is activated only when the hand and object are sufficiently close. Notably, we initiate the thumb motion first, followed by the motion of the other four fingers. This sequence is adopted because the thumb, equipped with two actuators and having greater weight, moves more slowly compared to the other fingers. The wrist trajectory is generated via motion planning using the Rapidly-exploring Random Tree (RRT) algorithm, ensuring feasible and collision-free paths for robotic motion. Specifically, the RRT planner moves the hand near the object and then returns it afterward. Once the hand is fully within the camera’s view, manipulation is entirely driven by our policy. No additional hand-crafted control is used during the core grasp.

\begin{wraptable}{r}{0.5\textwidth}
\vspace{-15pt}
\centering
\resizebox{0.5\textwidth}{!}{
\begin{tabular}{lcc}
\toprule
\textbf{Method} & \textbf{Success Rate (\%, $\uparrow$)} & \textbf{Contact Ratio (\%, $\uparrow$)} \\  
\midrule
\textsc{Dexplore} (vision) & {52.7} & {48.2} \\
\midrule
\textsc{Dexplore} (state) & 87.7 & 69.3 \\
\bottomrule
\end{tabular}
}
\vspace{-3pt}
\caption{
Comparison of our vision-based policy with the state-based policy shows that it performs well in manipulating objects with only sparse reference and partial observations.
}
\label{tb:ablation}
\vspace{-12pt}
\end{wraptable}

\begin{table}[h!]
\centering
\begin{tabular}{lc}
\toprule
\textbf{Parameter} & \textbf{Value} \\
\midrule

Object Friction Scaling & \(\sim \mathcal{U}(0.7, 1.1)\) \\
Object Restitution Scaling& \(\sim \mathcal{U}(0.7, 1.1)\) \\
Object Density Scaling  & \(\sim \mathcal{U}(0.5, 2)\) \\
Object Shape Scaling  & \(\sim \mathcal{U}(0.8^3, 1)\) \\

Point Cloud Additive & \(\sim \mathcal{U}(0, 0.01)\) \\
\bottomrule
\end{tabular}
\vspace{0.5em}
\caption{Physics parameters controlled by automatic domain randomization during learning progression.}
\end{table}

\noindent\textbf{Compute cost.} At inference time, the policy runs on an RTX 4090 paired with an Intel i9-14900K, with a total latency of approximately 80 ms per step. Of this, around 60–80 ms is spent on descriptor computation, which includes acquiring a depth image, masking out the background and robot arm via forward kinematics.